\documentclass[10pt,twocolumn,letterpaper]{article}

\usepackage[pagenumbers]{iccv} % To force page numbers, e.g. for an arXiv version

\definecolor{iccvblue}{rgb}{0.21,0.49,0.74}
\usepackage[pagebackref,breaklinks,colorlinks,allcolors=iccvblue]{hyperref}
\usepackage{enumitem} % enumitem 패키지 추가
\usepackage{indentfirst} % 패키지 추가
\usepackage{graphicx}
\usepackage{float}
\usepackage{bm}
\usepackage{multirow}
\usepackage{booktabs}
\usepackage{amssymb}
\usepackage{array}
\usepackage{mathtools}
\usepackage{comment}
\usepackage{colortbl} % 테이블 색상 지원 패키지
\usepackage{xcolor}   % 색상 
\DeclareMathOperator*{\argmax}{argmax} 

\usepackage[hypcap=false]{caption}

\def\paperID{9575} % *** Enter the Paper ID here
\def\confName{ICCV}
\def\confYear{2025}

\title{Text Embedding Knows How to Quantize Text-Guided Diffusion Models}

\author{
Hongjae Lee \hspace{.5cm} Myungjun Son \hspace{.5cm} Dongjea Kang\hspace{.5cm} Seung-Won Jung \thanks{corresponding author}\\
{Korea University} \\
{\tt\small \{jimmy9704,  sonbill,  kedjk11,  swjung83\}@korea.ac.kr} \\
}

\begin{document}
\maketitle
\begin{abstract}

Despite the success of diffusion models in image generation tasks such as text-to-image, the enormous computational complexity of diffusion models limits their use in resource-constrained environments. To address this, network quantization has emerged as a promising solution for designing efficient diffusion models. However, existing diffusion model quantization methods do not consider input conditions, such as text prompts, as an essential source of information for quantization. In this paper, we propose a novel quantization method dubbed Quantization of Language-to-Image diffusion models using text Prompts (QLIP). QLIP leverages text prompts to guide the selection of bit precision for every layer at each time step. In addition, QLIP can be seamlessly integrated into existing quantization methods to enhance quantization efficiency. Our extensive experiments demonstrate the effectiveness of QLIP in reducing computational complexity and improving the quality of the generated images across various datasets. The source code is available at our project page \url{https://github.com/jimmy9704/QLIP}.
\end{abstract}    
\section{Introduction}
\label{sec:intro}
Diffusion models have achieved remarkable success in various generative tasks, particularly in text-to-image generation~\cite{rombach2022high,ruiz2023dreambooth,zhang2023adding}. However, the huge computational complexity of diffusion models restricts their practical applicability, especially in resource-constrained environments. Specifically, diffusion models typically require hundreds of denoising steps to produce high-quality samples, making them considerably slower than generative adversarial networks (GANs)~\cite{goodfellow2014generative}. To this end, many studies introduced advanced training-free samplers to reduce the number of denoising iterations~\cite{song2020denoising,lu2022dpm,liu2022pseudo,bao2022analytic}. However, the decrease in iterations alone is insufficient since the denoising models used in each iteration possess billions of parameters~\cite{podell2023sdxl, saharia2022photorealistic, betker2023improving}, leading to increased computational complexity and high memory consumption.

Besides other approaches toward light-weight network designs, such as network architecture search~\cite{Liu_2018_ECCV, chu2023mixpath}, network pruning~\cite{han2015learning, sui2021chip}, and knowledge distillation~\cite{li2023automated, yang2023online}, network quantization has received particular attention due to its simplicity, the increasing demand for low-precision operations, and related hardware support~\cite{wang2019haq,dong2019hawq,zhang2023text}. Although various network quantization methods have been proposed for convolutional neural networks (CNNs), the direct use of such methods for diffusion models has shown significant performance degradation~\cite{shang2023post}. As tailored solutions for diffusion models, several methods have been proposed considering the time step of diffusion models for quantization. For example, PTQ4DM~\cite{shang2023post} performed 8-bit quantization by constructing a calibration dataset based on the time step. Q-Diffusion~\cite{li2023q} introduced an advanced method based on BRECQ~\cite{li2021brecq} and performed the performance analysis of quantization methods for text-guided diffusion models. In addition, TDQ~\cite{so2024temporal} performed quantization with varying activation scaling based on the diffusion time step. However, these methods do not use the text prompt provided to the diffusion model as valuable information for quantization, leaving room for further improvement.

\begin{figure*}[!t]
\centering
\includegraphics[width=.97\linewidth]{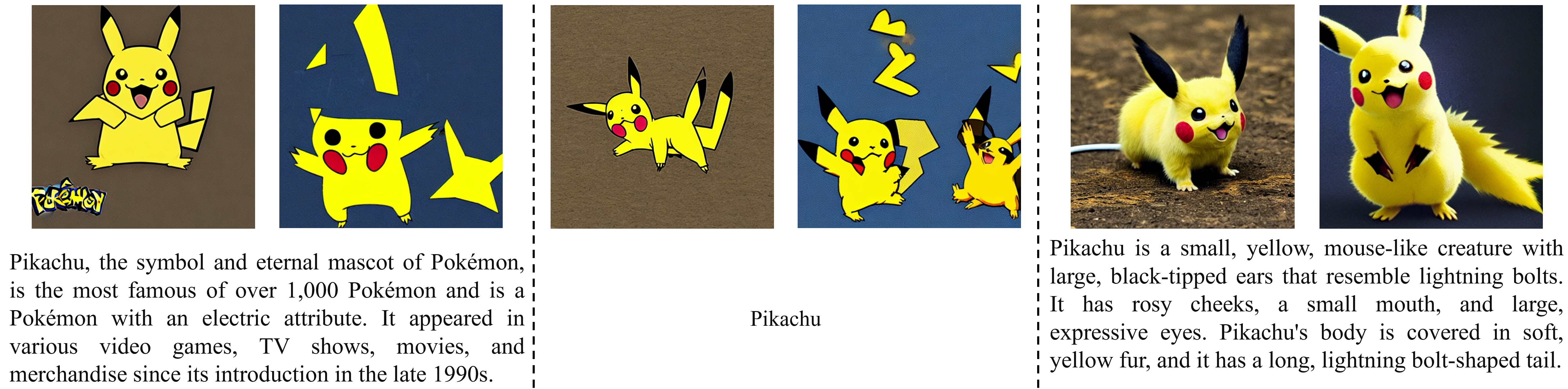}
\caption{Examples of the images generated by Stable Diffusion v1.4 using the text prompts given below images. When the text prompt details the shape, texture, and properties for the target object, the generated images have rich details and textures, as shown on the right; whereas the generated images are not very different from the images generated using only the keyword when the text prompt is less specific to the target object, as shown on the left.}
\label{fig:fig1}
\end{figure*}

Meanwhile, recent studies highlight the advantages of dynamic bit precision quantization considering input conditions ~\cite{liu2022instance, tian2023cabm}. For example, in the case of image super-resolution, CADyQ~\cite{hong2022cadyq} and AdaBM~\cite{hong2024adabm} performed dynamic quantization according to the complexity of the images and the importance of layers. RefQSR~\cite{lee2024refqsr} clustered the image patches and assigned high-bit and low-bit precisions to the reference and query patches, respectively. These dynamic quantization methods allow for the cost-effective assignment of bit precision, enhancing quantization efficiency. However, such input-adaptive quantization methods have not yet been explored in diffusion models.

In this paper, we propose a novel quantization method, dubbed Quantization of Language-to-Image diffusion models using text Prompts (QLIP), which determines the bit precision of diffusion models according to the text prompt. \Cref{fig:fig1} shows our observation: The quality of the generated images increases as the text prompt has more specific information about the image to be generated. This suggests that high-bit precision operations are necessary to generate images with vivid details and textures corresponding to detailed text prompts; whereas, low-bit precision operations can be sufficient to generate images corresponding to text prompts that lack specific guidance on details and textures.

\begin{figure*}[!t]
\centering
\includegraphics[width=.97\linewidth]{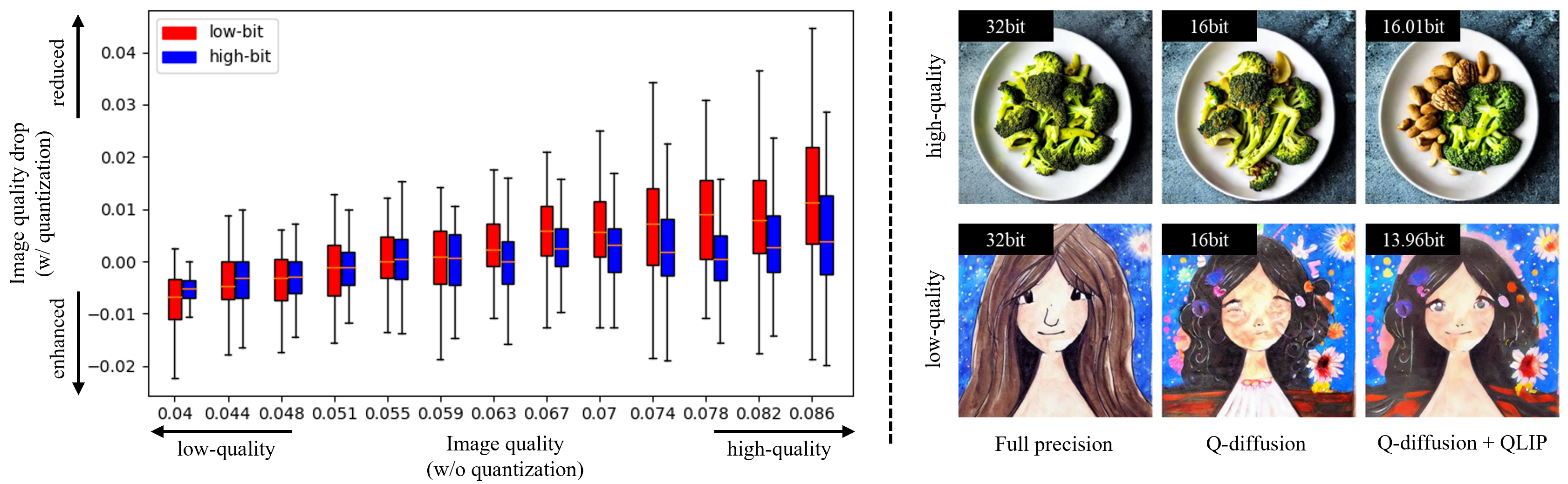}
\caption{Analysis of the effect of quantization and the quality of generated images, where the image quality is measured using the GIQA score~\cite{gu2020giqa}. On the left side, the x-axis represents the quality of images generated by the full-precision model, and the y-axis represents the quality difference between the generated images using the full-precision and quantized models. The images are generated by Stable Diffusion v1.4 ~\cite{rombach2022high} using 5000 captions from the COCO 2017 dataset~\cite{lin2014microsoft} with a size of 512$\times$512. The high-bit (W4A10) and low-bit (W4A6) quantized models are obtained by applying Q-diffusion~\cite{li2023q}. On the right side, examples of images generated using models that are unquantized or quantized w/wo QLIP are shown for two different image quality levels.}
\label{fig:fig2}
\end{figure*}

Motivated by this observation, we quantized the diffusion model~\cite{rombach2022high} using Q-diffusion~\cite{li2023q} with high-bit and low-bit settings and compared the quality of the generated images with the full-precision model. As shown in \Cref{fig:fig2}, the performance drop of the low-bit model is noticeable when the quality of the generated images is high for the full-precision model. In contrast, when the quality is low for the full-precision model, the high-bit and low-bit models perform similarly to the full-precision model. These results suggest that diffusion models can be efficiently quantized without sacrificing performance if the quality of the generated images can be first predicted and then used to determine bit precision. Accordingly, our proposed QLIP predicts the image quality from the text prompt and determines the quantization bit precision for diffusion models. 

Furthermore, previous studies~\cite{shang2023post,li2023q,so2024temporal,he2024ptqd} revealed that the activation distributions of diffusion models vary across layers over time steps. Inspired by these studies, QLIP uses learnable parameters to determine the bit precision for each layer and for each timestep. Our contributions can be summarized as follows:
\begin{itemize}
\item {To the best of our knowledge, QLIP is the first approach that explicitly utilizes text prompts for dynamic quantization in diffusion models.}
\item {We propose leveraging text prompts to predict the image quality, which is used to guide the selection of bit precision for every layer in different denoising time steps.}
\item {Our proposed QLIP can be integrated into existing diffusion quantization methods to enhance quantization efficiency.}
\end{itemize}

\section{Related work}
\subsection{Computationally efficient diffusion models}
Diffusion models generate high-quality images by iteratively removing noise from random noise. Despite their capability to generate high-quality images, diffusion models encounter challenges in real-world applications due to the time-consuming iterative denoising process. Thus, many efforts have been made to make the diffusion process computationally efficient, including reducing the number of iterative diffusion processes via a non-Markovian process~\cite{song2020denoising}, adjusting the variance scheduling~\cite{nichol2021improved}, and employing knowledge distillation~\cite{salimans2022progressive, meng2023distillation}. %For example, [progressive distill] used a denoising model trained with many sampling steps as a teacher to fine-tune a student denoising model with fewer sampling steps.

Meanwhile, since the computational complexity of diffusion model architectures also limits their practical applicability, especially in resource-constrained environments, research efforts have been made to lighten denoising models. For example, SnapFusion~\cite{li2024snapfusion} introduced an efficient U-Net architecture based on Stable Diffusion~\cite{rombach2022high}, while BK-SDM~\cite{kim2023bk} further lightened the architecture through knowledge distillation.

\subsection{Quantization of diffusion models}

Quantization involves mapping 32-bit floating-point values of weights and activations to lower-bit values. Numerous studies have explored quantization to compress and accelerate neural networks~\cite{choi2018pact, hong2022cadyq}, which can be classified into two approaches: quantization-aware training (QAT) and post-training quantization (PTQ). Recent studies~\cite{chu2024qncd, pandey2023softmax, ryu2024memory, tang2024pcr} on quantizing diffusion models have focused mainly on PTQ due to the inability to use training data for privacy and commercial reasons, as well as the high computational cost of QAT. Specifically, PTQ4DM~\cite{shang2023post} and Q-diffusion~\cite{li2023q} performed quantization of diffusion models by generating a time step-aware calibration set that consists of data sampled across various time steps using the full-precision denoising model. PTQD~\cite{he2024ptqd} separated the quantization noise from the diffusion noise to reduce quantization errors. Additionally, PTQD introduced a time step-based mixed-precision method but did not employ varying bit precision for each layer. Furthermore, TDQ~\cite{so2024temporal} performed quantization by adaptively scaling the activations for different diffusion time steps and layers, but assigned the same bit precision for all time steps. Although these diffusion model quantization methods have shown progress in reducing the complexity of noise estimation models, they do not adjust quantization parameters based on input conditions such as text prompts.

\subsection{Input-adaptive dynamic quantization}
Recent advances in hardware have enabled mixed-precision operations, allowing various layers to use different bit precisions to improve performance and computational efficiency~\cite{wang2019haq, dong2019hawq}. Many methods have been proposed to determine the bit precision of each layer of CNNs~\cite{wang2019haq,dong2019hawq,liu2022instance}. For example, several input-adaptive dynamic quantization methods ~\cite{hong2022cadyq, tian2023cabm, lee2024refqsr, hong2024adabm} have been introduced for super-resolution, which adjust bit precisions based on the complexity of image patches and the sensitivity to quantization. However, diffusion models have not yet explored such input-adaptive quantization methods. Therefore, this paper proposes a novel quantization method for text-guided diffusion models that applies adaptive quantization using input conditions.

\begin{figure*}[!t]
\centering
\includegraphics[width=.93\linewidth]{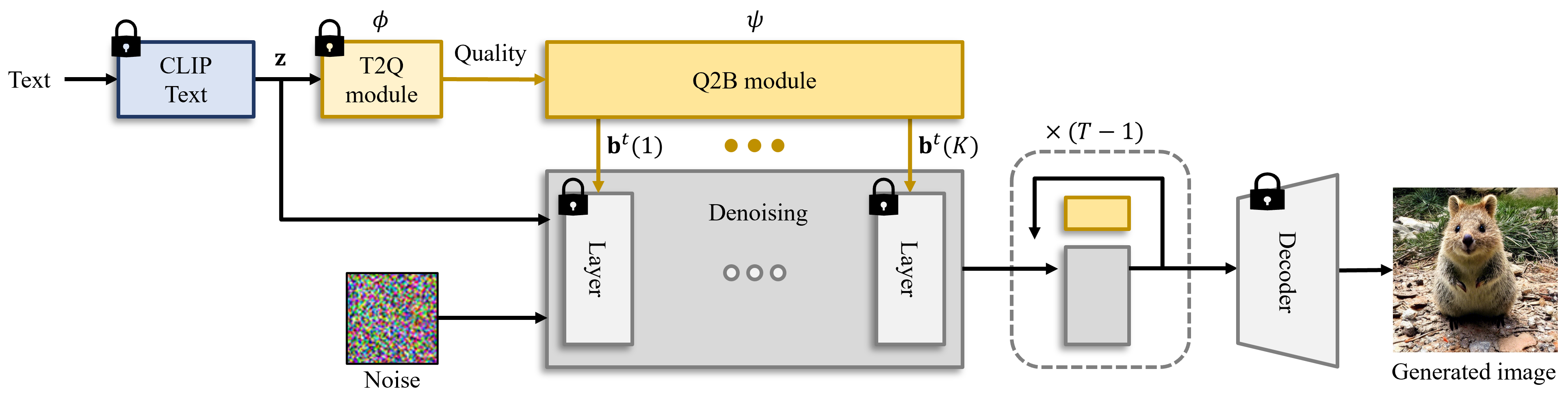}
\caption{Overview of QLIP that consists of two key modules: T2Q module that predicts image quality from the input text prompt and Q2B module that determines bit precisions for all denoising layers and time steps. Only the Q2B module is trained during QLIP training. See \Cref{sec:T2Q,sec:Q2B} for the details about the pre-training of the T2Q module and diffusion models.}
\label{fig:overview}
\end{figure*}
\section{Method}
\subsection{Preliminaries}
\textbf{Diffusion model.} The reverse process of a diffusion model gradually removes noise from random data $\mathbf{x}_t$ to produce a high-quality image $\mathbf{x}_0$, which can be expressed as $p_{\theta}(\mathbf{x}_{t-1}|\mathbf{x}_t) = \mathcal{N}(\mathbf{x}_{t-1}; \mu_{\theta}(\mathbf{x}_t, t), \sigma_{t}^2 \mathbf{I})$. The denoising process at time step $t$ using the noise estimation model $\epsilon_\theta$ can then be represented as follows:
\begin{equation}
\label{eq: reverse}
\mu_\theta(\mathbf{x}_t, t)  = \frac{1}{\sqrt{\alpha_t}}\left( \mathbf{x}_t - \frac{\beta_t}{\sqrt{1-\bar\alpha_t}} \epsilon_\theta(\mathbf{x}_t, t) \right) ,
\end{equation}
where $\alpha_t,\bar\alpha_t$, and $\beta_t$ are the parameteres that control the strength of the Gaussian noise in each step, and $\sigma_t$ is a variance schedule~\cite{ho2020denoising}.

\noindent \textbf{Diffusion model quantization.}
Diffusion model quantization reduces the complexity of a noise estimation model that requires hundreds of iterative sampling processes. The denoising process at time step $t$ using the quantized noise estimation model $\hat{\epsilon}_\theta$ and the quantized data $\hat{\mathbf{x}}_t$ can be expressed as follows:

\begin{equation}
\begin{aligned}
\label{eq: reverse quant}
\hat{\mu}_\theta(\hat{\mathbf{x}}_t, t)  = \frac{1}{\sqrt{\alpha_t}}\left( \hat{\mathbf{x}}_t - \frac{\beta_t}{\sqrt{1-\bar{\alpha}_t}} \hat{\epsilon}_\theta(\hat{\mathbf{x}}_t, t) \right),\\ \text{where} \; \hat{\epsilon}_\theta = Q_{\mathbf{b}^t}(\epsilon_\theta). 
\end{aligned}
\end{equation}     
Here, $\mathbf{b}^t \in\mathbb{R}^{K}$ represents a set of quantization bit precisions for $K$ layers at time step $t$, and $Q_{\mathbf{b}^t}$ is a function that performs layer-wise and time step-wise quantization using $\mathbf{b}^t$. To our knowledge, all previous diffusion quantization methods set the same bit precision to all layers, \ie, $\mathbf{b}^t(1)=\mathbf{b}^t(2)=\cdots=\mathbf{b}^t(K)$, and only a few methods~\cite{he2024ptqd, tang2024pcr} performed time step-wise dynamic quantization, \ie, $\mathbf{b}^{t_i}\ne\mathbf{b}^{t_j}$ for some time steps $t_i$ and $t_j$, $t_i\ne t_j$. However, such methods do not adapt the bit precision for each layer.
Our proposed QLIP determines $\mathbf{b}^t$ by training so that the quantization error can be effectively reduced without sacrificing the performance of diffusion models.

\subsection{Framework overview}

Our observations shown in \Cref{fig:fig2} indicate that bit precisions for quantization can be determined cost-effectively given the quality of the generated image. However, this image quality remains unknown during the inference stage, motivating us to predict it using an available input, \ie, text prompts. Specifically, if text prompts have rich and detailed information, the expected quality of the generated image is high, and vice versa. To this end, we introduce a text-to-quality (T2Q) module $\phi$ that predicts image quality $q$ from the embedding vector $\mathbf{z}$, \ie, $q = \phi\left(\mathbf{z}\right)$, where $\mathbf{z}$ is obtained by CLIP-Text~\cite{radford2021learning} using the input text prompt.

Next, as we discussed, high-bit quantization is required to generate images with vivid details and textures corresponding to detailed text prompts; while low-bit quantization is sufficient to generate images using less specific text prompts. Therefore, we design a quality-to-bit (Q2B) module $\psi$ that maps image quality $q$ to appropriate bit precision, \ie, $\mathbf{b}^t = \psi\left(q\right)$. Note that the Q2B module determines bit precisions for each time step for different layers since the activation distributions of diffusion models vary across layers over time steps.

In summary, the proposed QLIP obtains a quantized diffusion model $\hat{\epsilon}_\theta = Q_{\psi\left(\phi\left(\textbf{z}\right)\right)}(\epsilon_\theta)$ by learning the parameters of $\psi$ and $\phi$. The overview of the proposed QLIP is shown in \Cref{fig:overview}. The following subsections detail the T2Q and Q2B modules and their training procedure and loss functions. 

\subsection{Text-to-Quality estimation}\label{sec:T2Q}
The T2Q module, denoted as $\phi$, aims to predict the image quality from the input text prompt. $\phi$ consists of three linear layers and outputs an image quality score $q$ as $q=\phi\left(\textbf{z}\right)$, where $\textbf{z} \in\mathbb{R}^{C_{clip}}$ represents the CLIP text embedding, and $C_{{clip}}$ denotes its channel dimension. For the training of the T2Q module, we collect a dataset $\mathcal{T}_{T2Q}=\{\mathbf{z}^i, \mathbf{x}_0^i\}_{i=1}^{N_{T2Q}}$, where $\mathbf{z}^i$ and $\mathbf{x}_0^i$ are the $i$-th sample of the CLIP text embedding and its corresponding generated image obtained using the full-precision diffusion model, respectively, and $N_{T2Q}$ is the number of images in $\mathcal{T}_{T2Q}$. Let $\bar{q}^i$ represent the (pseudo) ground-truth image quality of $\mathbf{x}_0^i$, where we used GIQA~\cite{gu2020giqa} as the default quality measure, with an additional normalization to make the range of quality values [0, 1]. See \Cref{sec:ablation} for the performance analysis with different quality measures. The T2Q module is trained using the loss $L_{t2q}$:
\begin{equation}
\label{eq: loss t2q}
    L_{t2q}= \frac{1}{N_{T2Q}}\sum_{i=1}^{N_{T2Q}}\left ( {\bar{q}^i}-\phi\left ( \mathbf{z}^i  \right ) \right )^2.
\end{equation}

Once the T2Q module is trained, it is frozen during the training of the Q2B module and served as a quality predictor to help determine quantization bit precisions. 

\subsection{Quality-to-Bit estimation}\label{sec:Q2B}
In text-guided diffusion models, the richer and more detailed the information in the text prompt is, the higher the image quality tends to be, requiring the use of higher bits. Therefore, the image quality $q$ predicted by the T2Q module can serve as a direct hint for bit selection. To this end, the proposed Q2B module obtains a probability vector $\textbf{p}_q \in\mathbb{R}^{K}$ for bit selection as follows:
\begin{equation}
\label{eq: qbs}
    \textbf{p}_q = \sigma\left((q - 0.5)\textbf{s}  + \textbf{o}\right),
\end{equation}
where $\textbf{s} \in\mathbb{R}^{K}$ and $\textbf{o} \in\mathbb{R}^{K}$ represent sets of learnable parameters, and $\sigma$ denotes the sigmoid function. Furthermore, to adjust the bit precision for each time step, additional learnable parameters $\textbf{u}_m^t \in\mathbb{R}^{K}$ and $\textbf{u}_h^t \in\mathbb{R}^{K}$ are introduced to obtain probability vectors $\textbf{p}_m^t \in\mathbb{R}^{K}$ and $\textbf{p}_h^t \in\mathbb{R}^{K}$ as
\begin{equation}
\label{eq: tbs}
    \textbf{p}_{j}^{t} = \sigma(\textbf{u}_{j}^{t}), \;\; j \in \{m,h\}, t = 1, 2, \cdots, T,
\end{equation}
where $T$ is the total number of denoising time steps. 

These probability vectors, $\textbf{p}_q$, $\textbf{p}_m^t$, and $\textbf{p}_h^t$, are used to determine the bit precision for each layer and each time step. Specifically, we define a set of supported bit precisions as $\mathcal{B} = \{ b_{low}, b_{med}, b_{high}\}$. Then, the probabilities of selecting $b_{low}$, $b_{med}$, and $b_{high}$ are obtained as 
\begin{equation}
\begin{aligned}
\label{eq: prob}
    \textbf{p}_{b_{low}}^t &= (1-\textbf{p}_q) \odot (1-\textbf{p}_m^t),\\
    \textbf{p}_{b_{med}}^t &= (1-\textbf{p}_q) \odot \textbf{p}_m^t + \textbf{p}_q \odot (1-\textbf{p}_h^t), \\
    \textbf{p}_{b_{high}}^t &= \textbf{p}_q \odot \textbf{p}_h^t, 
\end{aligned} 
\end{equation} 
where $\odot$ represents element-wise multiplication, $\textbf{p}_{b_{low}}^t \in\mathbb{R}^{K}$, $\textbf{p}_{b_{med}}^t \in\mathbb{R}^{K}$, and $\textbf{p}_{b_{high}}^t \in\mathbb{R}^{K}$ are the probability vectors for bit selection. Finally, the bit precision for each layer at time step $t$ is determined as follows:
\begin{equation}
    \label{eqn: bit selection}
    \textbf{b}^t\left(k\right) = \argmax_{i \in \{b_{low},b_{med},b_{high}\}} \textbf{p}_i^t\left(k\right), ~k = 1, 2, \cdots, K.
\end{equation}

However, the training of $\textbf{u}_m^t$ and $\textbf{u}_h^t$ for all $t$ poses a challenge as $T$ increases. In addition, the activation distributions of the diffusion models are known to be similar to each other for adjacent time steps~\cite{he2024ptqd, so2024temporal}. Consequently, we learn the parameters for every $M$ time step, \ie, $\textbf{u}_{j}^{Mt}$, and use the same parameters for the intermediate time steps, \ie, $\textbf{u}_{j}^{M(t+1)-1} = \textbf{u}_{j}^{M(t+1)-2} = \cdots = \textbf{u}_{j}^{Mt}$, $j \in \{m,h\}$. Next, previous studies~\cite{balaji2022ediff, choi2022perception} revealed that the initial stage of the reverse process involves generating the rough context of an image corresponding to the input text prompt; thus, the use of low bits in the initial time steps could decrease the correlation between the text prompt and the generated image. Therefore, we set all values in $\textbf{p}_q$ to 1 for the first $m$ time steps in \Cref{eq: prob} to enforce the selection of high bits in the initial stage of the reverse process. In summary, the Q2B module has 2$K$ ($\textbf{s}$ and $\textbf{o}$) + 2$KT/M$ ($\textbf{u}_m^t$ and $\textbf{u}_h^t$) trainable parameters. 

The bit selection in \Cref{eqn: bit selection} is non-differentiable, making it unable to backpropagate. Thus, we adopt a straight-through estimator~\cite{bengio2013estimating} to render this process differentiable:
\begin{equation}
\begin{aligned}
    \label{eqn:method-bitloss4}
    &\hat{\textbf{a}}^t(k) = \\
    &\begin{cases}
    Q_{\textbf{b}^t(k)}(\textbf{a}^t(k)) & \text{forward}, \\
    \sum_{i \in \{b_{low},b_{med},b_{high} \}} \textbf{p}_i^t(k) \cdot Q_{i}(\textbf{a}^t(k)) & \text{backward}.
    \end{cases}
\end{aligned} 
\end{equation} 
Here, $\textbf{a}^t(k)$ and $\hat{\textbf{a}}^t(k)$ represent the activations of the full-precision and quantized models at time step $t$ of the $k$-th layer. In other words, during backpropagation, the activations quantized by all bit precisions in $\mathcal{B}$ are weighted by their corresponding estimated probabilities. Since diffusion models are less sensitive to the quantization of the weights~\cite{li2023q, he2024ptqd}, the weights are quantized using a fixed precision, which is empirically chosen as 4 bits for all layers and time steps.

Given the full precision diffusion model, we obtain its quantized weights by applying the baseline diffusion quantization methods~\cite{li2023q, he2024ptqd}. The quantization parameters, including the maximum and minimum values for clipping, are also predetermined for all bit precisions in $\mathcal{B}$. Then, while freezing all the other modules in \Cref{fig:overview}, only the Q2B module is trained using a small calibration set. The Q2B module is trained at randomly sampled time steps $t$ using the loss function $L_{QLIP}$:

\begin{equation}
\begin{aligned}
\label{eq: loss final}
    &L_{QLIP} = \left(\epsilon_\theta(\mathbf{x}_t, t) -\hat{\epsilon}_\theta(\hat{\mathbf{x}}_t, t)\right)^2 \\ 
    &+ \lambda_{bit}\left(b_{high} \cdot\sum_{k=1}^K\textbf{p}^t_{b_{high}}(k) +b_{med}\cdot\sum_{k=1}^K\textbf{p}_{b_{med}}^t(k)\right).
\end{aligned}
\end{equation}
Here, the first term measures the difference between the results of the full-precision and quantized models, and the second term measures the bit lengths for the high- and medium-precision models. $\lambda_{bit}$ is a weighting parameter that balances between these two terms. %The training of the Q2B module is performed using a small calibration set generated from the full-precision model.
\section{Experiments}
\subsection{Experimental setup}

\begin{table*}[!t]
  \aboverulesep=0.0ex
  \belowrulesep=0.0ex
  \renewcommand{\tabcolsep}{.5cm}
  \centering
  \begin{tabular}{llccccc}
  \midrule \midrule
    \multicolumn{7}{c}{COCO2017} \\
    \toprule
    \multicolumn{1}{c}{Method}     & \multicolumn{1}{c}{Bitoptions} &FAB$_\downarrow$ & BitOPs (T)$_\downarrow$ & FID$_\downarrow$ & sFID$_\downarrow$ & CLIP Score$_\uparrow$\\
    \midrule
    BK-SDM~\cite{kim2023bk}            & W32A32           & 32.00      & 10.46      & 23.79      & 66.19      & 0.3069\\
    \midrule
    Q-diffusion~\cite{li2023q}               & W4A16            &16.00       & 1.03       & 30.02      & 73.25      & \bf{0.3068} \\
    \rowcolor{gray!10}
    \multicolumn{1}{r}{+QLIP} & W4A\{8,16,32\}   &\bf{12.14}  & \bf{0.88}  & \bf{30.01} & \bf{73.24} & 0.3063 \\
    \midrule
    PTQD~\cite{he2024ptqd}                      & W4A16            &16.00       & 1.03       & 30.27       & 77.18     & \bf{0.3069}\\
    \rowcolor{gray!10}
    \multicolumn{1}{r}{+QLIP} & W4A\{8,16,32\}   &\bf{12.14}   & \bf{0.88} & \bf{30.02} & \bf{73.26} & 0.3063 \\
    \midrule \midrule
    \multicolumn{7}{c}{Conceptual Captions} \\
    \toprule 
    \multicolumn{1}{c}{Method}     & \multicolumn{1}{c}{Bitoptions} &FAB$_\downarrow$ & BitOPs (T)$_\downarrow$ & FID$_\downarrow$ & sFID$_\downarrow$ & CLIP Score$_\uparrow$\\
    \midrule
    BK-SDM~\cite{kim2023bk}           & W32A32           & 32.00      & 10.46      & 20.00           & 47.80      & 0.2983 \\
    \midrule
    Q-diffusion~\cite{li2023q}               & W4A16            & 16.00      & 1.03      & \bf{24.68}       & 62.70      & \bf{0.2965} \\
    \rowcolor{gray!10}
    \multicolumn{1}{r}{+QLIP} & W4A\{8,16,32\}   & \bf{10.58} & \bf{0.82}      & 24.72       & \bf{59.54} & 0.2964 \\
    \midrule
    PTQD~\cite{he2024ptqd}   & W4A16            & 16.00      & 1.03      & \bf{24.66}       & 62.60      & 0.2965 \\
    \rowcolor{gray!10}
    \multicolumn{1}{r}{+QLIP} & W4A\{8,16,32\}   & \bf{10.58} & \bf{0.82}      & 24.71       & \bf{59.25} & \bf{0.2966} \\
    \bottomrule
  \end{tabular}
  \caption{Quantitative comparisons of various quantization methods at a resolution of 512$\times$512 using BK-SDM-Tiny-2M.}
  \label{tab: BK}
\end{table*}

\begin{table}[!th]
  \aboverulesep=0.0ex
  \belowrulesep=0.0ex
  \renewcommand{\tabcolsep}{.1cm}
  \centering
  \begin{tabular}{lcccc}
    \midrule \midrule
    \multicolumn{5}{c}{COCO2017} \\
    \toprule
    \multicolumn{1}{c}{Method}  &FAB$_\downarrow$ & FID$_\downarrow$ & sFID$_\downarrow$ & CLIP Score$_\uparrow$\\
    \midrule
    Stable Diffusion~\cite{rombach2022high}&  32.00       & 22.23      & 65.11      & 0.3174\\
    \midrule
    Q-diffusion~\cite{li2023q}&  8.00       & 23.40      & 66.57      & \bf{0.3126}\\
    \rowcolor{gray!10}
    \multicolumn{1}{r}{+QLIP} &  \bf{7.86}  & \bf{21.61} & \bf{64.32} & 0.3120\\
    \midrule
    PTQD~\cite{he2024ptqd}    &  8.00       & 22.75      & 68.63      & \bf{0.3126}\\
    \rowcolor{gray!10}
    \multicolumn{1}{r}{+QLIP} &  \bf{7.86}  & \bf{21.35} & \bf{65.81} & 0.3120\\
    \midrule \midrule
    \multicolumn{5}{c}{Conceptual Captions} \\
    \toprule
    \multicolumn{1}{c}{Method}&FAB$_\downarrow$ & FID$_\downarrow$ & sFID$_\downarrow$ & CLIP Score$_\uparrow$\\
    \midrule
    Stable Diffusion~\cite{rombach2022high}&  32.00       & 15.67      & 45.61      & 0.3078\\
    \midrule
    Q-diffusion~\cite{li2023q}&  8.00       & 20.74      & 48.33      & \bf{0.2984}\\
    \rowcolor{gray!10}
    \multicolumn{1}{r}{+QLIP} &  \bf{7.77}  & \bf{19.89} & \bf{47.63} & 0.2979\\
    \midrule
    PTQD~\cite{he2024ptqd}    &  8.00       & 25.40      & 55.01      & 0.2972\\
    \rowcolor{gray!10}
    \multicolumn{1}{r}{+QLIP} &  \bf{7.77}  & \bf{20.42} & \bf{48.83} & \bf{0.2975}\\
    
    \bottomrule
  \end{tabular}
  \caption{Quantitative comparisons of various quantization methods at a resolution of 512$\times$512 using Stable Diffusion v1.4. For the bit precision options, W4A8 is used for Q-diffusion and PTQD, and W4A\{6,8,10\} are used for QLIP.}
  \label{tab: SDM}
\end{table}

\textbf{Datasets and baseline quantization methods.} To evaluate the effectiveness of QLIP across different diffusion models, we conducted experiments using two diffusion networks:  BK-SDM-Tiny-2M~\cite{kim2023bk} and Stable Diffusion v1.4~\cite{rombach2022high}. Our experiments involve two datasets, COCO2017~\cite{lin2014microsoft} and Conceptual Captions~\cite{sharma2018conceptual}. To train the T2Q module, 10k images were generated by Stable Diffusion v1.4~\cite{rombach2022high} using 10k text annotations in the COCO2017 training dataset as input text prompts. We generated another 1024 images using unused text annotations in the COCO2017 training dataset to train the Q2B module. We evaluated the effectiveness of our proposed QLIP by applying it to the baseline diffusion quantization methods, Q-diffusion~\cite{li2023q} and PTQD~\cite{he2024ptqd}. In the implementation of PTQD, we set the noise correction parameter to $b_{med}$, while noise correction was not applied if the estimated image quality by the T2Q module is below 0.3 or above 0.7. We trained the T2Q module for 3 epochs using the Adam optimizer~\cite{kingma2014adam} with a learning rate of 0.001 and the Q2B module for 5000 iterations using BRECQ~\cite{li2021brecq}, as adopted in Q-diffusion and PTQD, using RTX 3090.
For performance evaluation, following the baseline quantization method~\cite{li2023q}, we generated 10k images using 10k text annotations in the COCO2017 validation dataset. Moreover, to evaluate the generalization ability of the quantized models, we used Conceptual Captions to generate additional 10k images. 

\noindent \textbf{Evaluation metrics.} We report widely adopted metrics such as FID~\cite{heusel2017gans} and sFID~\cite{NIPS2016_sfid} to evaluate image quality. In addition, we use CLIP Score~\cite{hessel2021clipscore} to assess the degree of matching between the generated images and their corresponding text prompts. To quantify computational efficiency, we measure BitOPs~\cite{van2020bayesian, lee2024refqsr}, which account for the required operations in both the denoising model and additional modules for QLIP implementation, and feature average bit-width (FAB)~\cite{hong2024adabm, hong2022cadyq} for the denoising model. In particular, BitOPs are measured as $MACs \cdot \frac{b_w}{32} \cdot \frac{b_a}{32}$, where $MACs$ denotes multiply-accumulate computations, and $b_w$ and $b_a$ represent the bitwidths of weights and activations, respectively.

\begin{figure}[!t]
\centering
\includegraphics[width=1.\linewidth]{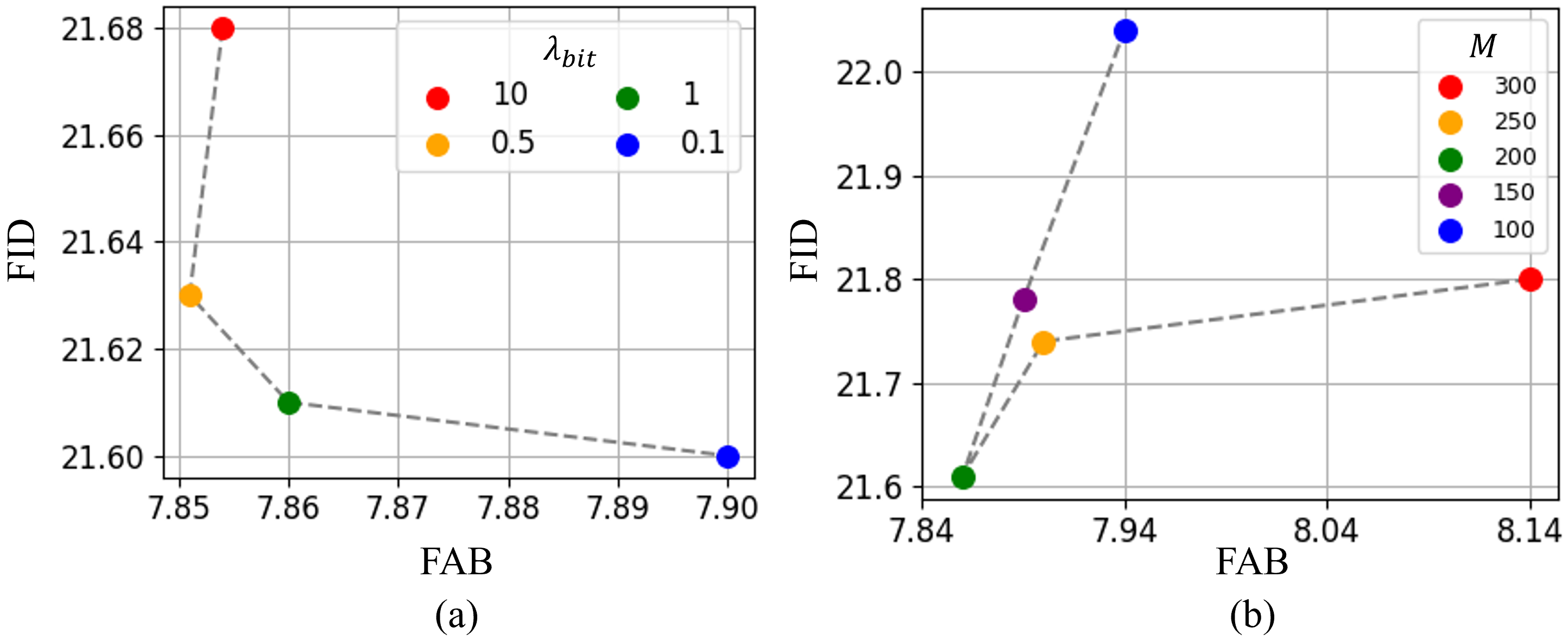}
\caption{Results on Q-diffusion using different hyperparameter settings of QLIP (W4A\{6,8,10\}): (a) $\lambda_{bit}$ and (b) $M$.}
\label{fig: hyper}
\end{figure}

\noindent \textbf{Hyperparameter settings.} We conducted experiments on various hyperparameter settings for the COCO2017 validation dataset using Q-diffusion as the baseline in Stable Diffusion v1.4. First, \Cref{fig: hyper}(a) presents the results obtained with various values of the weight term $\lambda_{bit}$ in \Cref{eq: loss final}, where $\lambda_{bit}$ = 1 showed an effective balance between FID and FAB. Next, \Cref{fig: hyper}(b) shows the results for different $M$ values in $\textbf{u}_{j}^{Mt}, j \in \{m,h\}$ when the maximum timestep $T$ was 1000, where we obtained the best scores with $M$ = 200.  

\begin{table}[!t]
    \centering
    \aboverulesep=0.0ex
    \belowrulesep=0.0ex
    \renewcommand{\tabcolsep}{.05cm}
    \begin{tabular}{llccc}
      \toprule
      Method                   &Bitoptions      &Runtime (s)$_\downarrow$ & FAB$_\downarrow$ & FID$_\downarrow$ \\
      \midrule
      BK-SDM~\cite{kim2023bk}  &W32A32          &  6.50          & 32.00            & 20.00 \\
      \midrule
      Q-diffusion~\cite{li2023q}&W4A8            &  \textbf{4.53} & \textbf{8.00}    & 28.32 \\
      Q-diffusion~\cite{li2023q}&W4A16           &  5.58          & 16.00            & \textbf{24.68} \\
      \rowcolor{gray!10}
      \multicolumn{1}{r}{+QLIP}&W4A\{8,16,32\}  &  4.85  & 10.58 & 24.72 \\
      \bottomrule
    \end{tabular}
    \caption{Runtime comparisons, where W4A\{8,16,32\} is used for QLIP. Measured on the Conceptual Captions dataset.}
    \label{tab:runtime}
\end{table}

\begin{figure*}[!t]
\centering
\includegraphics[width=.89\linewidth]{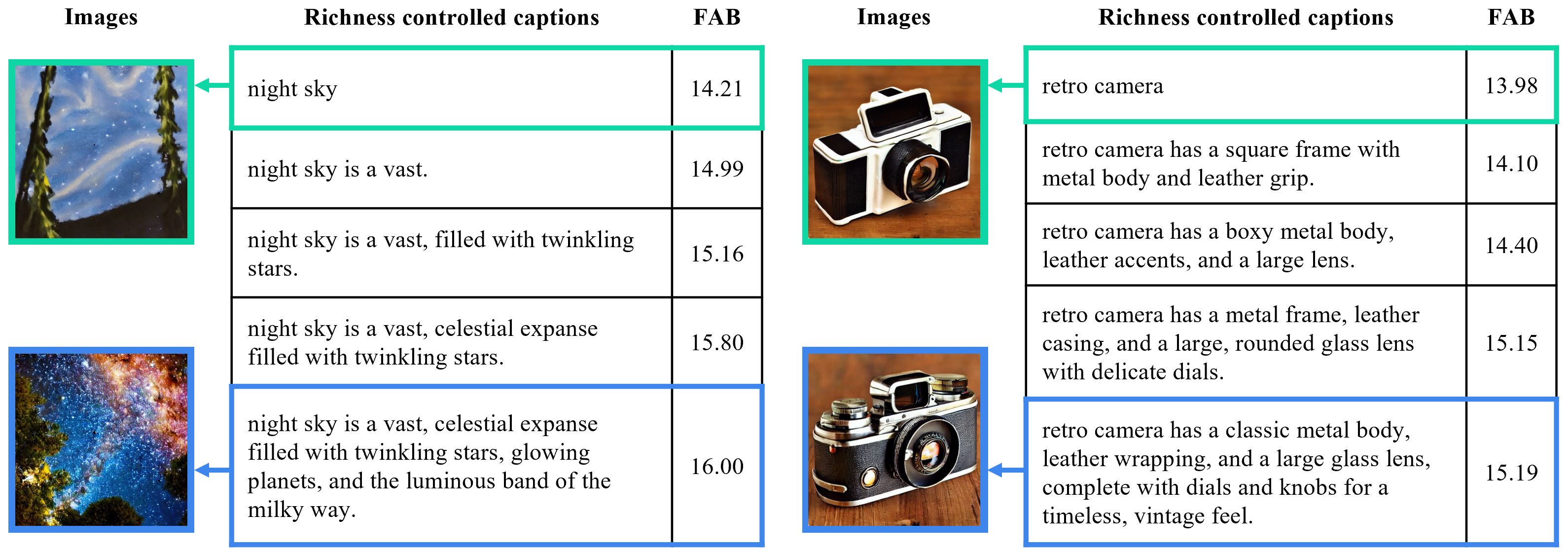}
\caption{Examples of variations in FAB by QLIP for the texts with different levels of richness and detail, along with the generated images.}
\label{fig: controlled_captions_main}
\end{figure*}

\begin{figure*}[!t]
\centering
\includegraphics[width=.89\linewidth]{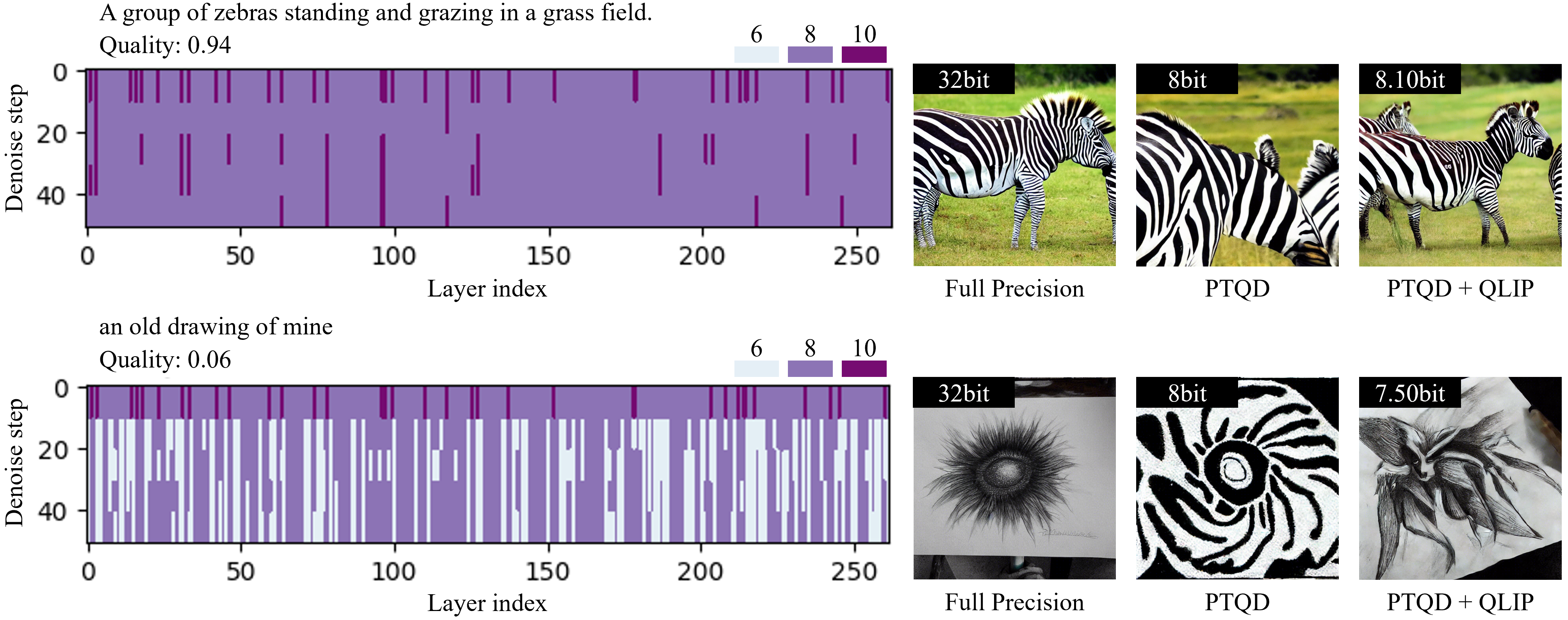}
\caption{Examples of the bit selection results and generated images. When the text prompt is specifically describing the image to be generated, the predicted image quality is high, and high and medium bits are assigned for most layers and time steps, as shown on the top left. On the other hand, when the text prompt has less detailed information, the predicted image quality is low, and low bits are assigned for many layers and time steps, as shown on the bottom left. On the right side, the generated images using Q-diffusion with and without applying QLIP are shown, demonstrating the effectiveness of QLIP in saving bits while improving the quality of generated images.}
\label{fig: main_result}
\end{figure*}

\subsection{Quantitative results}
\noindent \textbf{Comparison with existing quantization methods.}
To demonstrate the effectiveness and applicability of QLIP, we tested QLIP on two diffusion model quantization methods: Q-diffusion~\cite{li2023q} and PTQD~\cite{he2024ptqd}. \Cref{tab: BK,tab: SDM} provides comparison results for the COCO2017~\cite{lin2014microsoft} and Conceptual Captions~\cite{sharma2018conceptual} datasets using lightweight diffusion model BK-SDM-Tiny-2M~\cite{kim2023bk} and Stable Diffusion v1.4~\cite{rombach2022high}, respectively. 

As shown in \Cref{tab: BK}, for the COCO2017 dataset, QLIP effectively reduced computational complexity of the baseline diffusion quantization methods. For example, Q-diffusion+QLIP (W4A\{8,16,32\}) achieved a reduction in FAB by 3.86 while maintaining similar FID and sFID, compared to Q-diffusion (W4A16). Furthermore, the use of QLIP consistently led to improvements when applied to PTQD, as demonstrated in \Cref{tab: BK}. For example, PTQD+QLIP (W4A\{8,16,32\}) enhanced FID and sFID by 0.25 and 3.92, respectively, while also reducing FAB and BitOPs. The results on Conceptual Captions followed trends similar to those observed in COCO2017. For example, Q-diffusion+QLIP (W4A\{8,16,32\}) demonstrated reduced computational complexity compared to Q-diffusion on Conceptual Captions.

\Cref{tab: SDM} presents the results of applying lower-bit precision combinations to Stable Diffusion v1.4. Here, PTQD+QLIP (W4A\{6,8,10\}) achieved a 4.98 improvement in FID compared to PTQD (W4A8), while maintaining similar FAB using bit combinations efficiently. Although our proposed QLIP drives the quantization methods to reduce computational complexity and improve image quality, the CLIP Scores remained similar to those of the baseline quantization methods, indicating that the fitness to the text prompts is not significantly influenced by QLIP. Additional results on larger diffusion models can be found in the supplementary material.

\noindent \textbf{Runtime measurement.}
Runtime was measured with Cutlass~\cite{cutlass}, including the T2Q and Q2B modules, as shown in \Cref{tab:runtime}. The results show that runtime was reduced to a level similar to W4A8 while maintaining an FID score comparable to W4A16. Measurements were conducted on an RTX 3090. Additional analysis on memory requirements can be found in the supplementary material.

\subsection{Qualitative results}\label{sec:qualitative}
\Cref{fig: controlled_captions_main} visualizes the changes in FAB by the richness and detail level of the text, along with the generated images. Here, ChatGPT-4o was used to create texts of varying complexity that describe a given keyword. Specifically, the prompt used was ``Create four sentences that describe how `keyword' looks like, gradually getting richer and longer.'' As can be seen, QLIP assigned higher bits as the text describes the keyword more specifically and in greater detail, demonstrating that it utilizes information from the text to determine bit precisions for effective synthesis of vivid details and textures.

\Cref{fig: main_result} shows several results comparing Q-diffusion with and without QLIP when applied to Stable Diffusion v1.4. As can be seen, our QLIP generates more detailed and natural images compared to PTQD by using various bit combinations. Furthermore, \Cref{fig: main_result} visualizes the bit precision maps determined by QLIP. When the T2Q module predicts high image quality from input text prompts, QLIP allocates more bits to enable rendering of vivid and accurate details from rich text prompts. Otherwise, QLIP employs lower bits for layers that are less sensitive to quantization, based on the text prompt and the time step. For example, when given less specific text prompts, the correlation between the generated image and the text prompt is relatively less critical, allowing for low-bit quantization in the cross-attention blocks. These results highlight the effectiveness and significance of dynamically allocating bit precisions according to text prompts.

\subsection{Ablation studies}\label{sec:ablation}
\textbf{Ablation studies on the T2Q module.} We report the results on the COCO2017 validation dataset using Stable Diffusion v1.4 with Q-diffusion as the baseline quantization method. \Cref{tab: ab-t2q} presents the results of the T2Q modules trained using various perceptual image quality metrics, including Realism score~\cite{NEURIPS2019_realism}, CLIP-IQA~\cite{wang2023exploring}, and GIQA~\cite{gu2020giqa}. The performance of the T2Q module was evaluated using Spearman’s rank-order correlation coefficient (SROCC)~\cite{spearman1961proof} and Pearson’s linear correlation coefficient (PLCC)~\cite{pearson1895notes} between the actual quality measures and the predictions of the T2Q module. QLIP with GIQA showed the most substantial improvements in FAB and FID along with the highest SROCC and PLCC scores. %Moreover, computational complexity, for example FAB, uniformly decreased across all metrics.

\noindent \textbf{Ablation studies on the Q2B module.} We conducted experiments on several variants of the Q2B module using Q-diffusion and Stable Diffusion v1.4 on the COCO2017 validation dataset. As shown in \Cref{tab: ab-q2b}, when using either $\textbf{p}_q$ alone or in combination with $\textbf{p}_h^t$, effective reduction in FAB was achieved at the expense of increase in FID. When using $\textbf{p}_q$ and $\textbf{p}_m^t$, FID decreased while FAB increased. Finally, when $\textbf{p}_q$, $\textbf{p}_h^t$, and $\textbf{p}_m^t$ were all involved, balanced results were obtained in two metrics. Implementation details of the ablation studies on the Q2B module can be found in the supplementary material.
\section{Conclusion}

\begin{table}[!t]
  \aboverulesep=0.0ex
  \belowrulesep=0.0ex
  \renewcommand{\tabcolsep}{.15cm}
  \centering
  \begin{tabular}{llccccc}
    \toprule
    \multicolumn{1}{c}{Method} & SROCC$_\uparrow$ & PLCC$_\uparrow$ &FAB$_\downarrow$ & FID$_\downarrow$\\
    \midrule
    Q-diffusion~\cite{li2023q}           & - & -  & 8.00 & 23.40\\
    \midrule
    +QLIP  \\
    \quad w/ Realism score  & 0.5132 & 0.5022 & 8.10 & 22.18 \\
    \quad w/ CLIP-IQA  & 0.7127 & 0.7082 & 8.54 & 21.81 \\
    \quad w/ GIQA & \bf{0.8047} & \bf{0.8108} & \bf{7.86}  & \bf{21.61} \\
    \bottomrule
  \end{tabular}
  \caption{Ablation studies on the T2Q module: For the bit precision options, W4A8 in Q-diffusion and +QLIP W4A\{6,8,10\} were used.} 
  \label{tab: ab-t2q}
\end{table}

\begin{table}[!t]
  \aboverulesep=0.0ex
  \belowrulesep=0.0ex
  \renewcommand{\tabcolsep}{.1cm}
  \centering
  \begin{tabular}{llccccc}
    \toprule
    \multicolumn{1}{c}{Method}     & \multicolumn{1}{c}{Bitoptions} &FAB$_\downarrow$ & FID$_\downarrow$ \\
    \midrule
    Q-diffusion~\cite{li2023q}               & W4A8            & 8.00 &23.40 \\
    \midrule
    \quad + $\textbf{p}_q$  & W4A\{6,10\}  & 7.57 & 26.91\\
    \quad + $\textbf{p}_q+\textbf{p}_h^t$  & W4A\{6,8,10\} & \bf{6.73} & 29.37 \\
    \quad + $\textbf{p}_q+\textbf{p}_m^t$  & W4A\{6,8,10\} & 8.60 & 21.96\\
    \midrule
    \quad + $\textbf{p}_q+\textbf{p}_m^t+\textbf{p}_h^t$ (QLIP) & W4A\{6,8,10\} & 7.86 & \bf{21.61} \\
    \bottomrule
  \end{tabular}
  \caption{Ablation studies on the Q2B module.}
  \label{tab: ab-q2b}
\end{table}

The proposed quantization method for diffusion models, called QLIP, is the first method that uses text prompts for quantization in diffusion models. From our observation that the bit precision for quantization can be effectively determined based on the quality of the generated images, we designed the T2Q module that predicts image quality from input text prompts. We then introduced the Q2B module that allocates appropriate bit precision for each time step for different layers. By integrating the T2Q and Q2B modules into existing diffusion model quantization methods, we demonstrated the effectiveness of the proposed QLIP. QLIP presents a promising solution for quantizing diffusion models due to its ability to improve the quality of generated images using text prompts while reducing computational costs.

\noindent \textbf{Limitations and future directions.} Similar to other generative models, QLIP also has the potential to generate counterfeit images for malicious purposes, potentially undermining trust in visual media. In a future study, we will extend QLIP to consider other input conditions, such as images or segmentation maps, in addition to text prompts.

\section*{Acknowledgement}
{This work was supported by the National Research Foundation of Korea (NRF) grant funded by the Korea government (MSIT) (No. RS-2022-NR070077).}

{
    \small
    \bibliographystyle{ieeenat_fullname}
    \bibliography{main}
}

\clearpage
\setcounter{page}{1}
\maketitlesupplementary

\setcounter{section}{0}
\renewcommand{\thesection}{\Alph{section}}

\setcounter{table}{0}
\renewcommand{\thetable}{S\arabic{table}}
\setcounter{figure}{0}
\renewcommand{\thefigure}{S\arabic{figure}}

In this supplementary document, we present additional results and analyses, including the following:
\begin{itemize}
    \item Results on a large-scale diffusion model (\Cref{sec: sdxl}).
    \item Evaluation on mixed-precision quantization method (\Cref{sec: pcr}).
    \item Ablation studies on bit selection criteria (\Cref{sec: crieria}).
    \item Ablation studies on bit precision options (\Cref{sec: bitoptions}).
    \item Batch inference scenario analysis (\Cref{sec: batch}).
    \item Memory requirement analysis (\Cref{sec: memory}).
    \item Details on the structure and implementation of the T2Q module (\Cref{sec: struct_t2q}).
    \item Implementation details for ablation studies (\Cref{sec: imp_ablation}).
    \item Quantization sensitivity analysis for each layer (\Cref{sec: sensitivity}).
    \item Additional visual comparison results.(\Cref{sec: add_fig}).
\end{itemize}

\section{Results on a Large-Scale Diffusion Model}\label{sec: sdxl}
We conducted experiments using SDXL~\cite{podell2023sdxl} with the Euler scheduler as the base model to evaluate the effectiveness of QLIP on a large-scale diffusion model. As shown in \Cref{tab: sdxl}, QLIP demonstrated significant improvements in computational efficiency while maintaining competitive image quality.

For the COCO2017 dataset, QLIP effectively reduced the computational complexity of the baseline diffusion quantization methods. Specifically, Q-diffusion+QLIP (W4A\{8,16,32\}) demonstrated a significant reduction in FAB while maintaining comparable FID and sFID scores to Q-diffusion (W4A16). This result indicates that QLIP optimizes the bit precision selection effectively, reducing computational overhead without compromising image quality. The results on the Conceptual Captions dataset exhibited similar trends to those observed with COCO2017. These results suggest that QLIP generalizes well to large-scale diffusion models.

\begin{table}[!t]
  \renewcommand{\tabcolsep}{.18cm}
  \centering
  \begin{tabular}{lcccc}
    \midrule \midrule
    \multicolumn{5}{c}{COCO2017} \\
    \toprule
    \multicolumn{1}{c}{Method}  &FAB$_\downarrow$ & FID$_\downarrow$ & sFID$_\downarrow$ & CLIP Score$_\uparrow$\\
    \midrule
    SDXL~\cite{podell2023sdxl}&  32.00       & 23.75      & 65.85      & 0.3180\\
    \midrule
    Q-diffusion~\cite{li2023q}               &  16.00       & 28.46      & 67.54      & 0.3178\\
    \rowcolor{gray!10}
    \multicolumn{1}{r}{+QLIP} & 12.68 &  28.16  & 66.31& 0.3177\\
    \midrule \midrule
    \multicolumn{5}{c}{Conceptual Captions} \\
    \toprule
    \multicolumn{1}{c}{Method}&FAB$_\downarrow$ & FID$_\downarrow$ & sFID$_\downarrow$ & CLIP Score$_\uparrow$\\
    \midrule
    SDXL~\cite{podell2023sdxl}&  32.00       & 19.25      & 47.72      & 0.3085\\
    \midrule
    Q-diffusion~\cite{li2023q}               &  16.00       & 22.23      & 49.57      & 0.3075\\
    \rowcolor{gray!10}
    \multicolumn{1}{r}{+QLIP} &  11.34  & 21.95 & 48.61 & 0.3074\\
    \bottomrule
  \end{tabular}
  \caption{Quantitative comparisons at a resolution of 768$\times$768 using SDXL~\cite{podell2023sdxl}. For the bit precision options, W4A16 was used for Q-diffusion and W4A\{8,16,32\} were used for QLIP.}
  \label{tab: sdxl}
\end{table}
  
 \begin{table}[!t]
  \renewcommand{\tabcolsep}{.25cm}
  \centering
  \begin{tabular}{l ccc}
  \toprule
    \multicolumn{1}{c}{Method} & FAB$_\downarrow$ & Image Reward$_\uparrow$ & Pick Score$_\uparrow$ \\
  \midrule
  FLUX~\cite{flux} & 16.00 & 1.1013 & 23.07 \\
  \midrule
  PCR~\cite{pcr} & 9.60 & 0.9986 & 22.97 \\
  \rowcolor{gray!10}
  \multicolumn{1}{r}{+QLIP} & 7.92 & 1.0214 & 23.01\\
  \bottomrule
\end{tabular}
  \caption{Quantitative comparisons with the mixed-precision quantization method PCR, using FLUX as the baseline model. Evaluation was conducted on 500 prompts from the COCO2017 dataset at a resolution of 1024$\times$1024. For the bit precision options, W4A\{8,16\} was used for PCR, and W4A\{6,8,16\} were used for QLIP.}
  \label{tab: flux}
\end{table}

\begin{table}[!t]
  \centering
  \renewcommand{\tabcolsep}{.5cm}
  \label{tab: crieria}
  \begin{tabular}{lcc}
    \toprule
    Bit Selection Strategy & FAB $\downarrow$ & FID $\downarrow$ \\
    \midrule
    Image Complexity                & 14.86          & 31.09 \\
    Prompt Length                   & 12.79          & 31.49 \\
    \midrule
    \rowcolor{gray!10}
    Image Quality (QLIP)            & 12.14 & 30.01 \\
    \bottomrule
  \end{tabular}
  \caption{Comparison of bit selection criteria.}
  \label{tab: crieria}
\end{table}

\section{Evaluation on Mixed-Precision Method} \label{sec: pcr}
We compared with PCR~\cite{pcr}, a recent mixed-precision quantization method, using FLUX~\cite{flux} as the base model. For image quality, we reported ImageReward~\cite{imagereward} and PickScore~\cite{pickscore}, using 500 prompts from COCO2017. As shown in \Cref{tab: flux}, applying our QLIP to PCR further improved performance by achieving higher quality scores while also reducing overall bit usage, demonstrating its effectiveness even on the recent diffusion model FLUX.

\section{Ablation Studies on Bit Selection Criteria} \label{sec: crieria}
We conducted experiments to explore alternative criteria for determining bit precision, replacing the predicted image quality used in our T2Q module. Specifically, we investigated two alternative metrics: image complexity and prompt length, as shown in Table~\ref{tab: crieria}. 

Image complexity has been effectively used in other tasks, such as super-resolution, as a criterion for dynamic quantization~\cite{hong2022cadyq}. To test its applicability in diffusion models, we replaced the T2Q and Q2B modules with T2C and C2B modules that utilize image complexity, measured as average image gradient magnitude. However, this configuration resulted in worse performance (FAB 14.86, FID 31.09) compared to the original QLIP design (FAB 12.14, FID 30.01), suggesting that image complexity alone is not a reliable criterion for bit selection in diffusion models.

We also evaluated a prompt length-based bit allocation strategy, where the number of tokens in the input prompt was used to decide bit precision. This variant also underperformed (FAB 12.79, FID 31.49) relative to our original approach. Unlike prompt length, which only reflects the input length, the T2Q module captures richer semantic representations from text, leading to more accurate bit assignment and better image quality.

\begin{table}[!t]
    \centering
    \renewcommand{\tabcolsep}{.75cm}
    \begin{tabular}{lcc}
      \toprule
      Bit-Options  & FAB$_\downarrow$ & FID$_\downarrow$ \\
      \midrule
      \{8,16\}      & 10.51 & 24.78 \\
      \{8,16,32\}   & 10.58 & 24.72 \\
      \{6,8,16,32\} & 9.24  & 25.22 \\
      \bottomrule
    \end{tabular}
    \caption{Ablation study on bit precision options for QLIP.}
    \label{tab:bit_options}
\end{table}

\begin{figure}[!t]
  \centering
    \includegraphics[width=.7\linewidth]{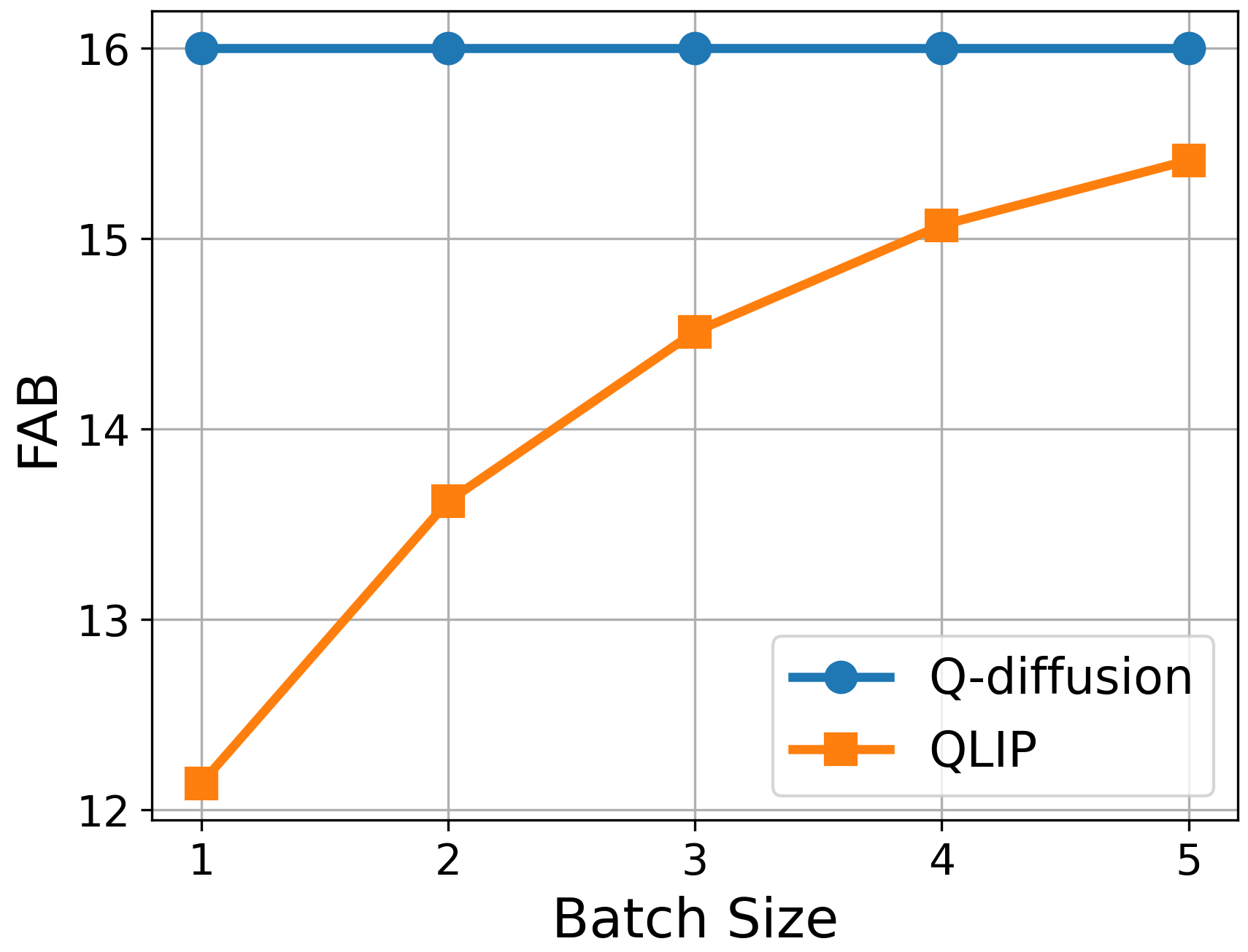}
    \caption{Bit allocation over different batch sizes.}
    \label{fig:batchsize}
\end{figure}

\begin{table}[!t]
\renewcommand{\tabcolsep}{.42cm}
    \centering
        \begin{tabular}{cccccc}
            \toprule
            \multirow{2}{*}{Layer No.} & \multirow{2}{*}{Operator} & Kernel \\
            &&($C_{in} \times C_{out}$) \\
            \midrule
            1 &  Linear / ReLU & $ C_{clip} \times C_{clip}$ \\
            2 &  Linear / ReLU & $ C_{clip} \times 512$ \\
            3 &  Linear / ReLU & $ 512 \times 1$ \\
            4 &  Sigmoid & - \\
            \bottomrule
        \end{tabular}
    \caption{The structure of the T2Q module.}
    \label{tab: t2q_arch}
\end{table}

\section{Ablation Studies on Bit-Options} \label{sec: bitoptions}
The impact of different bit-options on FAB and FID is presented in \Cref{tab:bit_options}. The results demonstrate that while the \{6,8,16,32\} configuration achieves the lowest FAB, it also leads to an increase in FID. Conversely, the \{8,16,32\} configuration results in the lowest FID while maintaining a relatively low FAB, making it a better choice for minimizing image degradation and effectively reducing computational overhead. Based on these findings, we select \{8,16,32\} as the default bit-option.

\begin{figure}[!t]
\centering
\includegraphics[width=.94\linewidth]{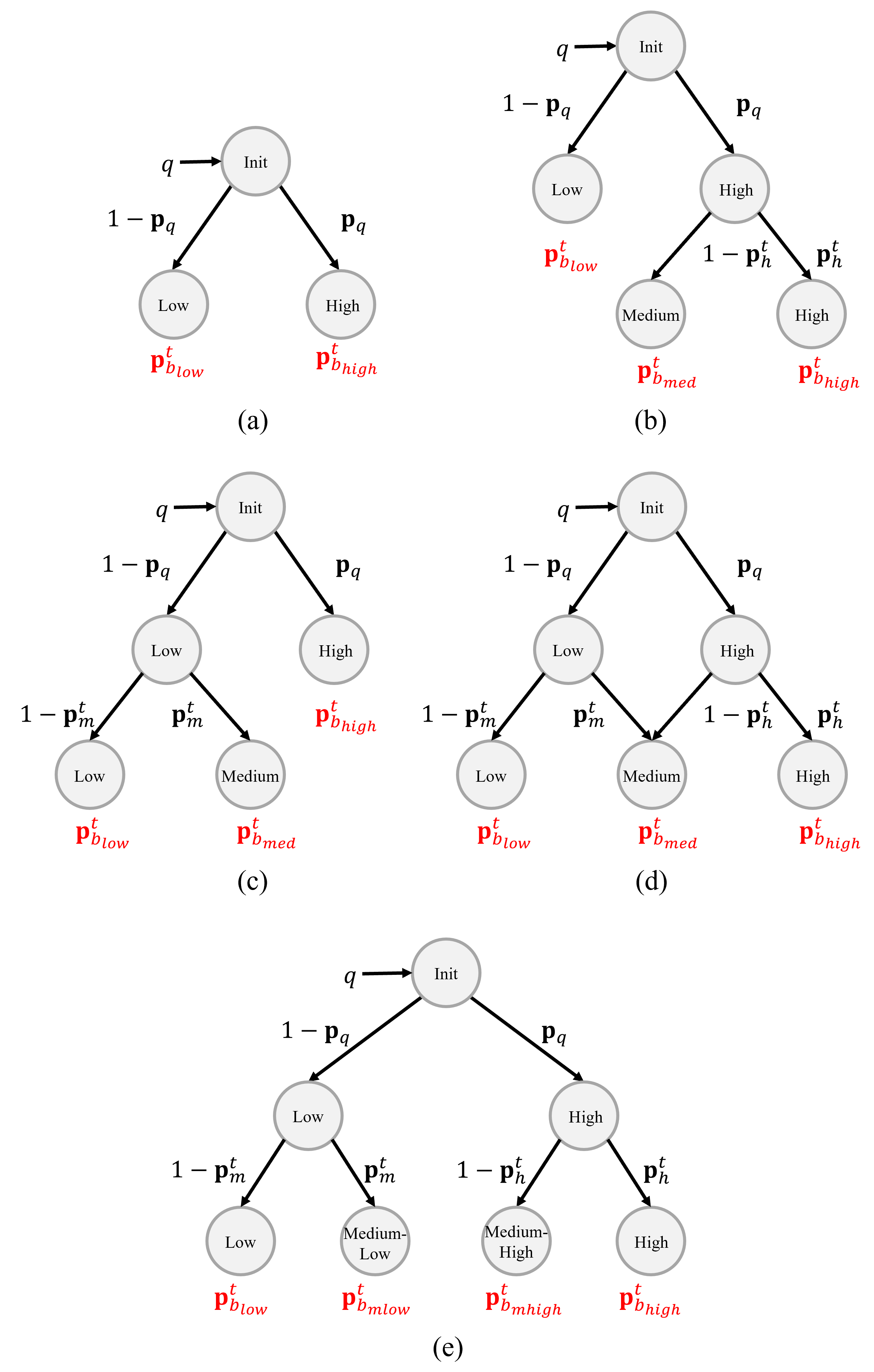}
\caption{Implementation details of ablation study variants for the Q2B module.
(a) Bit-Options correspond to the case with 2 bit candidates using $\textbf{p}_q$.
(b), (c), and (d) correspond to the case with 3 bit candidates using
$\textbf{p}_q + \textbf{p}_h^t$,
$\textbf{p}_q + \textbf{p}_m^t$, and
$\textbf{p}_q + \textbf{p}_m^t + \textbf{p}_h^t$, respectively.
(e) Bit-Options correspond to the case with 4 bit candidates.}
\label{fig: ablation}
\end{figure}

\section{Batch Inference Scenario} \label{sec: batch}
Our practical implementation strategy for batch processing is to assign the maximum required bit precision for each layer across the batch. As shown in \cref{fig:batchsize}, the effectiveness of QLIP decreases as the batch size increases, which is a known limitation of input-adaptive quantization methods. However, text-to-image generation is typically performed on individual prompts or small batches. For example, the recent GPT Image 1 API~\cite{gpt} provided by OpenAI officially supports small batch sizes per API call, and processing many images requires issuing multiple parallel API calls, rather than increasing the batch size within a single request. Therefore, this limitation has less significant practical impact.
  
\section{Memory Requirements} \label{sec: memory}
QLIP requires additional memory to store the model, including 1.5MB for the T2Q module and 12.1KB for the Q2B module, and additional 2.2KB for scale and zero-point values. However, this additional memory overhead is negligible compared to the total memory of BK-SDM-Tiny-2M 4bit (154.2MB). While our method does not reduce memory usage, it improves both energy efficiency and inference time by reducing the computational cost.

\section{Structure of the T2Q module }\label{sec: struct_t2q}

As shown in \Cref{tab: t2q_arch}, the T2Q module consists of a simple 3-layer MLP with a ReLU activation function. The first linear layer is the projection layer of the CLIP text encoder, which is frozen during training. In the final layer, a sigmoid function is used to limit the output range of the quality $q$. %, and subsequently, 0.5 is subtracted when computing $\textbf{p}_q$ to center it around zero. 

\section{Implementation Details on Ablation Studies}\label{sec: imp_ablation}
The implementation of the Q2B modules used in ablation studies is shown in \Cref{fig: ablation}. When only $\textbf{p}_q$ is used, $\textbf{p}^t_{med}$ is not included, limiting the bit-options to two candidates. When $\textbf{p}^t_l$ $\left(\textrm{or}~\textbf{p}^t_h\right)$ is not used, $\textbf{p}^t_{low} \left(\textrm{or}~\textbf{p}^t_{high}\right)$ is determined using $\textbf{p}_q$, adjusting the available three-bit candidates accordingly. Incorporating additional bit-options, such as $\textbf{p}^t_{mlow}$ and $\textbf{p}^t_{mhigh}$, can further increase the number of bit candidates.

\section{Analysis on Quantization Sensitivity} \label{sec: sensitivity}
\begin{figure*}[!ht]
\centering
\includegraphics[width=1.\linewidth]{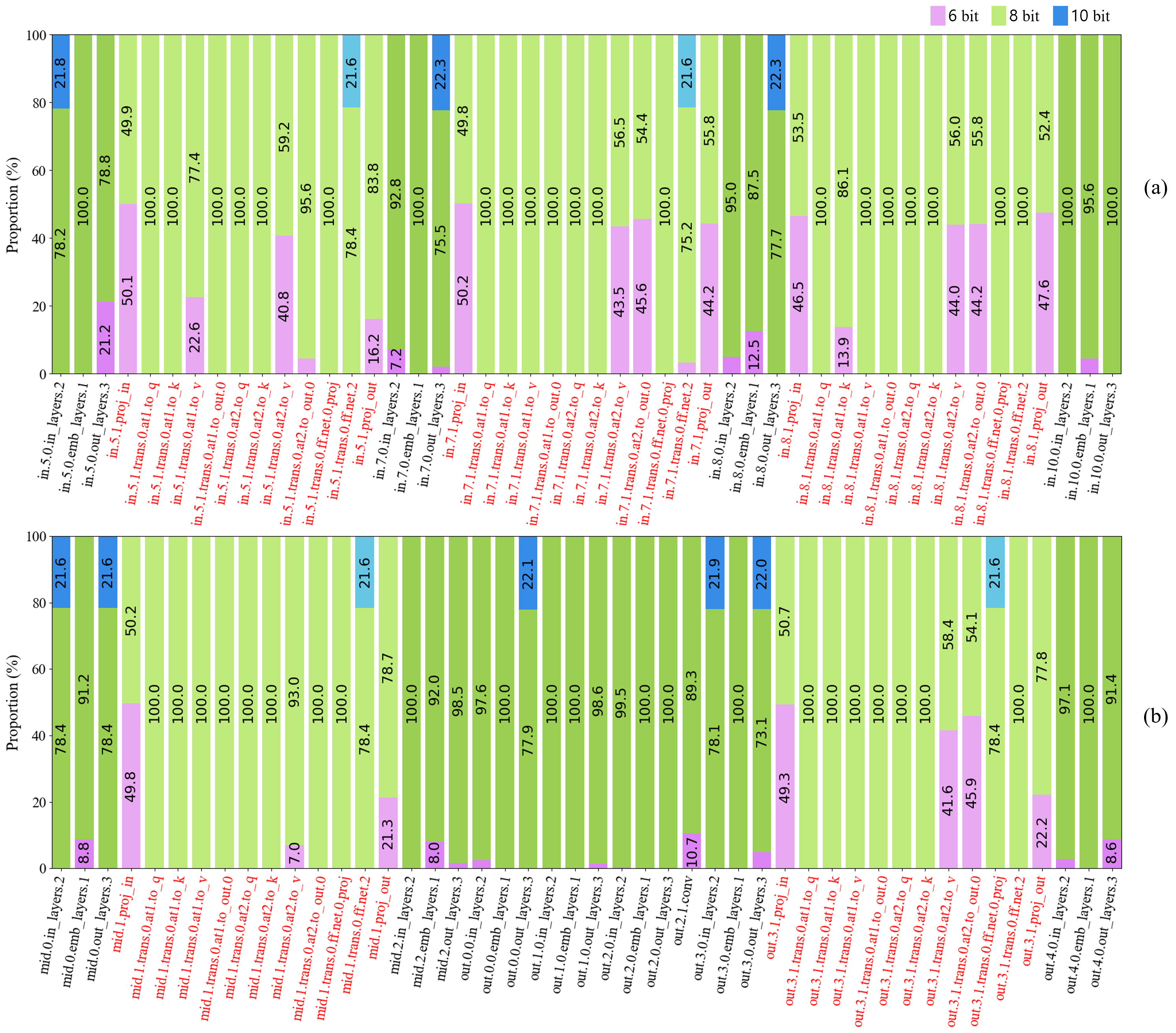}
\caption{The proportion of bits assigned in each layer of the Stable Diffusion model quantized by W4A\{6,8,10\} using Q-diffusion w/ QLIP while generating 10k images using the COCO2017 validation dataset. On the x-axis, cross-attention blocks and residual blocks are indicated in red and black, respectively. (a) and (b) show the statistics of several layers in the input, middle and output blocks.}
\label{fig: layer_bit}
\end{figure*}

\Cref{fig: layer_bit} shows the proportion of bits used in each layer of the quantized denoising model. Stable Diffusion utilizes cross-attention blocks to incorporate text prompts into the image latent space, in addition to basic residual blocks. Notably, our proposed QLIP aims to manage the overall bit usage, particularly by assigning low bits more frequently on the layers in cross-attention blocks. Specifically, it is evident that bit reduction occurs mainly within three layers: (1) “proj\_in” which leverages the denoised image latent passed into the cross-attention block, (2) “at2.to\_v” which projects text prompt to denoised image latent, (3) “proj\_out” which passes out the result of cross-attention to the next denoising blocks. These results are attributed to many cases with a weak correlation between the generated image and the text prompt, promoting the T2Q and Q2B modules to determine low bits for efficient quantization.

\section{Additional Qualitative Results}\label{sec: add_fig}
\begin{figure*}[!t]
\centering
\includegraphics[width=1.\linewidth]{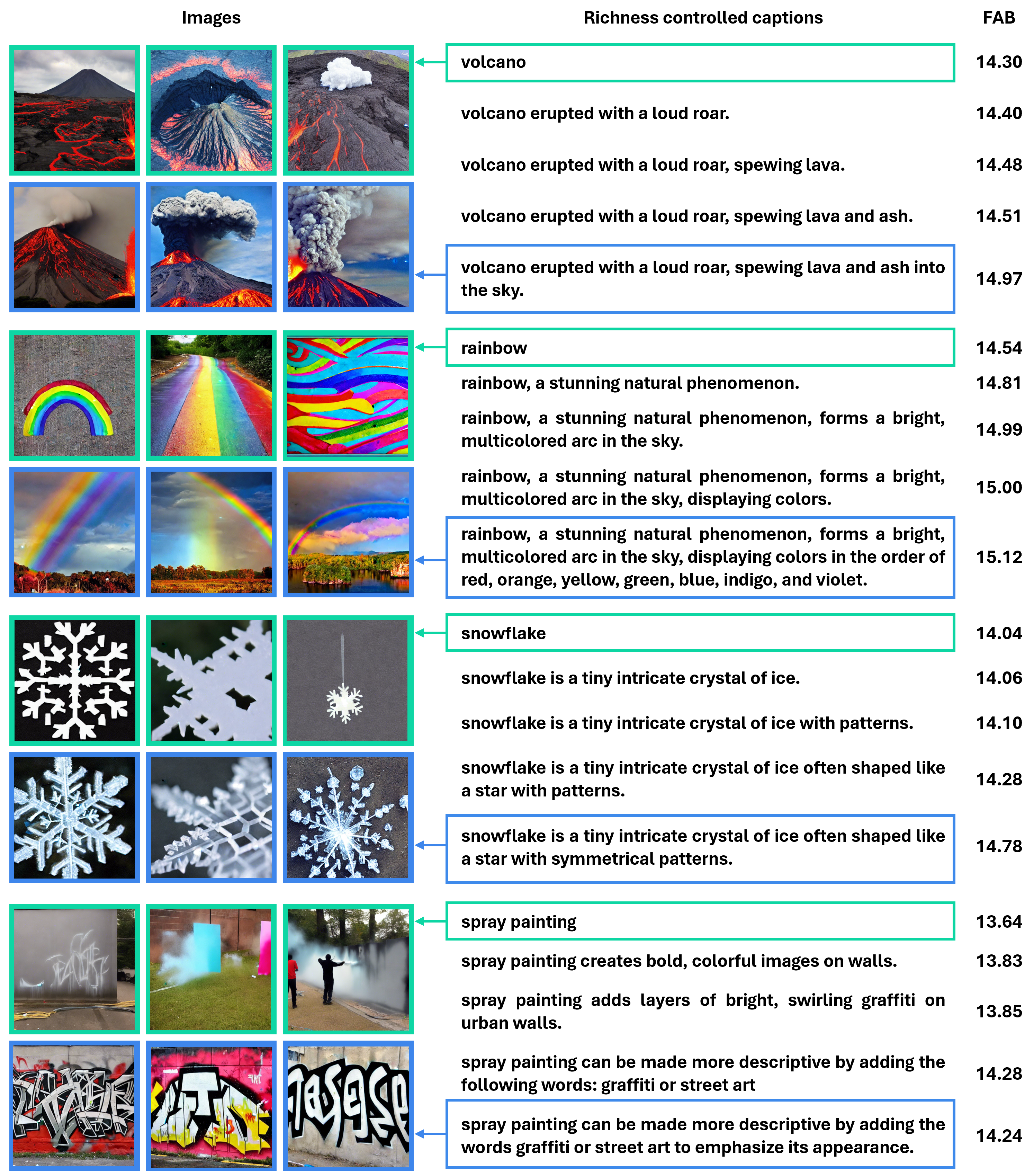}
\caption{Examples of variations in FAB by QLIP for the texts with different levels of richness and detail, along with the generated images.}
\label{fig: controlled_captions}
\end{figure*}

\Cref{fig: controlled_captions} presents additional examples illustrating how changes in FAB are influenced by the richness and specificity of the text descriptions, along with the generated images. As shown, QLIP assigned higher bits when the text provides more specific and detailed descriptions, demonstrating that it leverages textual information to adapt bit precision, enabling effective synthesis of vivid details and textures.

\begin{figure*}[!t]
\centering
\includegraphics[width=1.\linewidth]{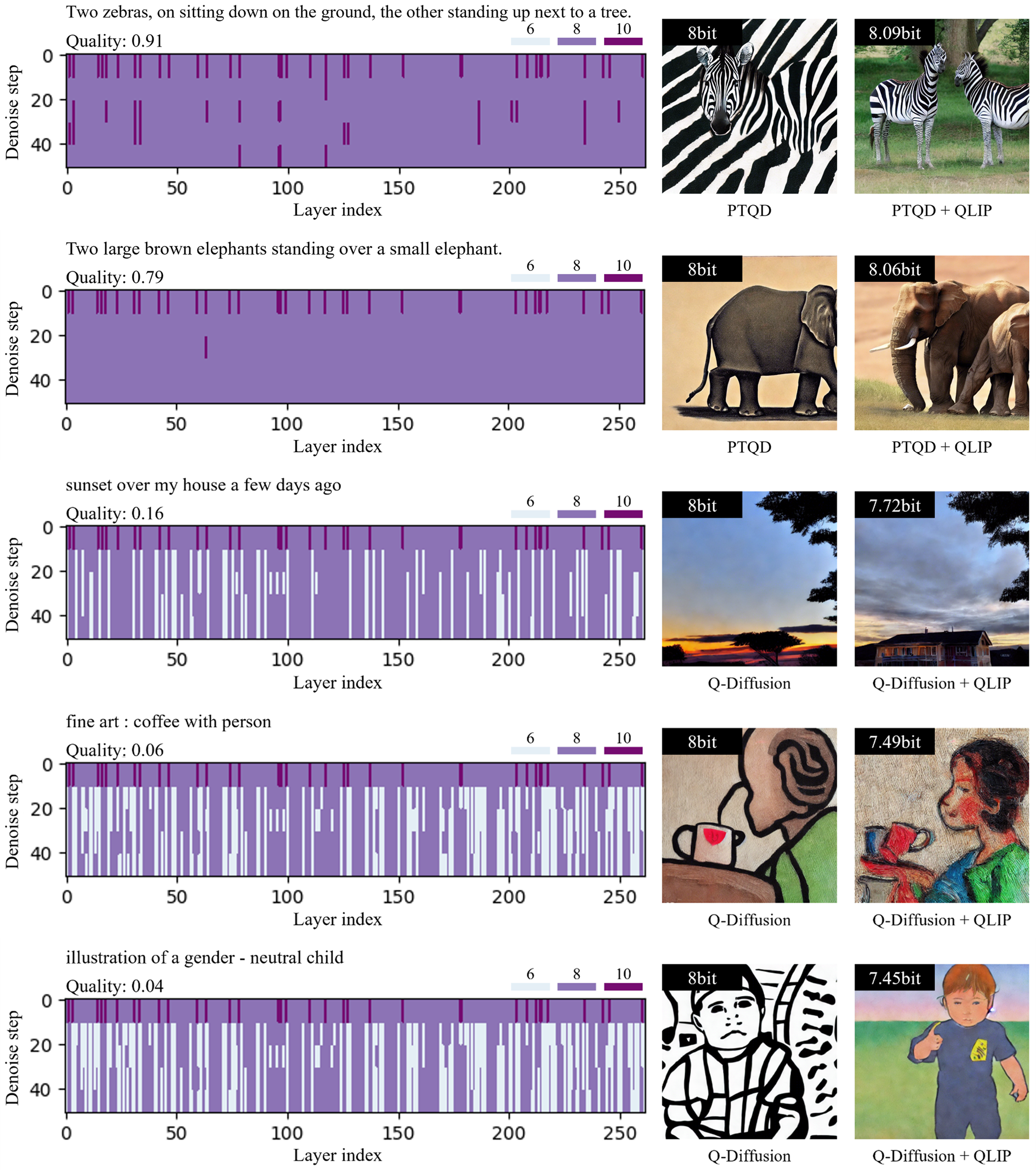}
\caption{Examples of the bit selection results and generated images using Q-diffusion or PTQD as baseline quantization methods. QLIP is applied with the bit precisions of W4A\{6,8,10\}.}
\label{fig: more_result_1}
\end{figure*}

\begin{figure*}[!t]
\centering
\includegraphics[width=1.\linewidth]{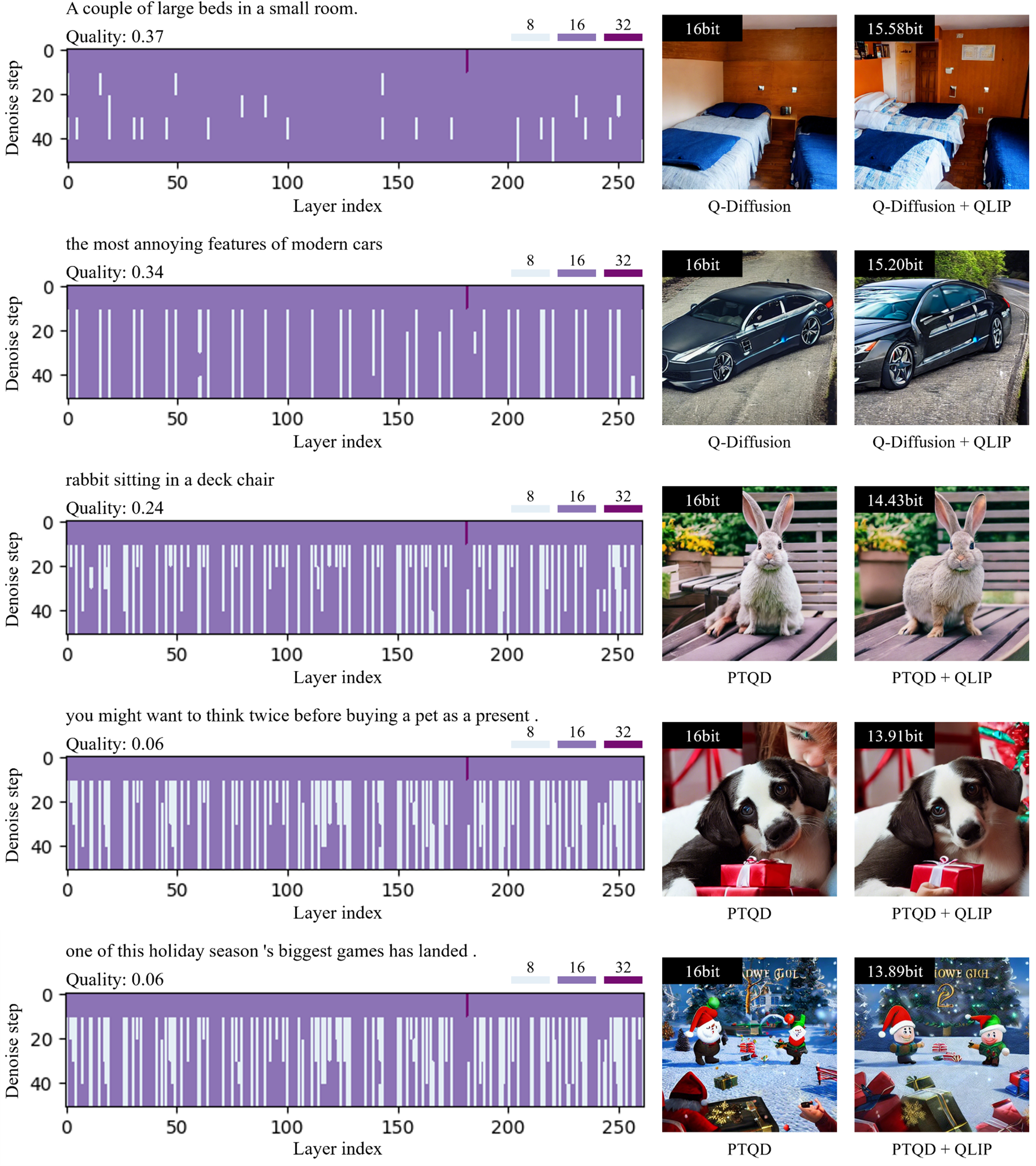}
\caption{Examples of the bit selection results and generated images using Q-diffusion or PTQD as baseline quantization methods. QLIP is applied with the bit precisions of W4A\{8,16,32\}.}
\label{fig: more_result_2}
\end{figure*}

\begin{figure*}[!t]
\centering
\includegraphics[width=1.\linewidth]{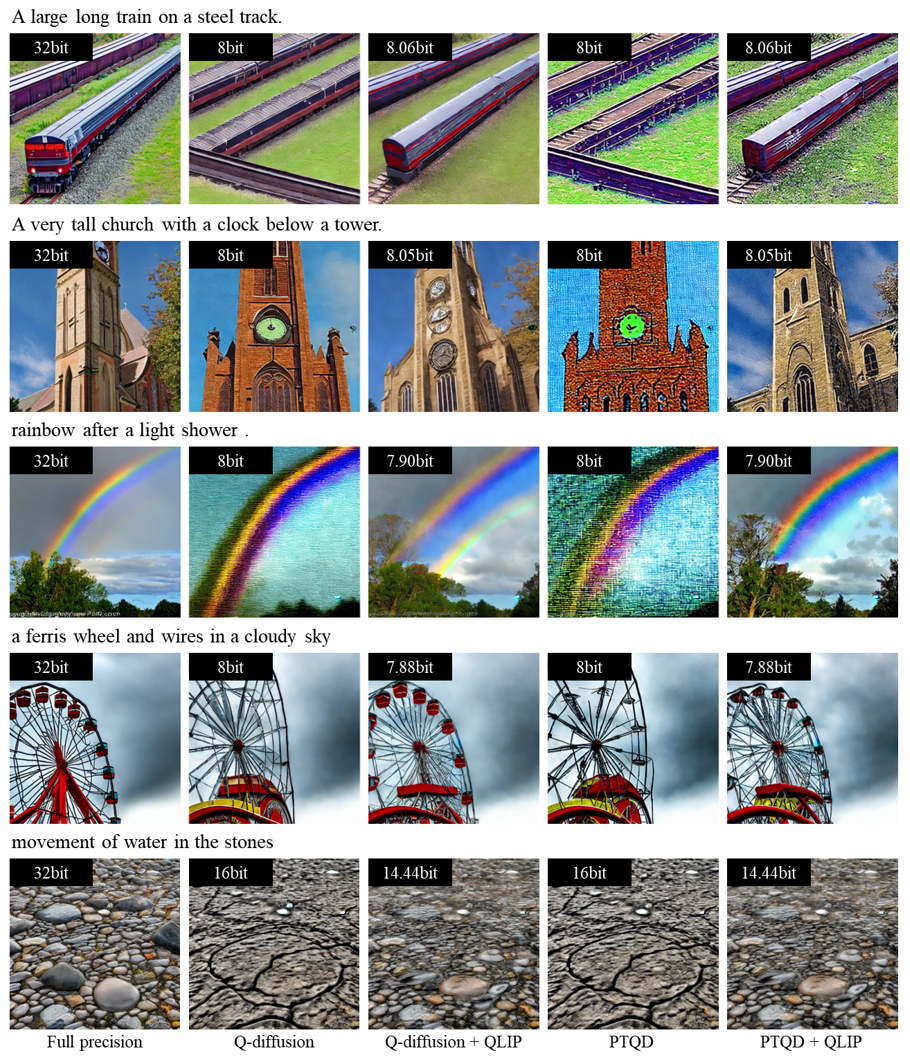}
\caption{Examples of generated images using QLIP with Q-diffusion or PTQD as baseline quantization method.}
\label{fig: more_result_3}
\end{figure*}

\Cref{fig: more_result_1,fig: more_result_2} show additional examples of the bit selection results and generated images using Q-diffusion or PTQD as baseline quantization methods. \Cref{fig: more_result_3} provides examples of images generated by QLIP using Q-diffusion or PTQD as baseline quantization methods, along with the full-precision model. The text prompts for generating images were sourced from the captions of COCO2017 and Conceptual Captions datasets.

\end{document}